\documentclass{article}

\usepackage[preprint]{neurips_2026}
\usepackage{graphicx}

\usepackage[utf8]{inputenc} 
\usepackage[T1]{fontenc}    
\usepackage{hyperref}       
\usepackage{url}            
\usepackage{booktabs}       
\usepackage{amsfonts}       
\usepackage{nicefrac}       
\usepackage{microtype}      
\usepackage{xcolor}         

\usepackage{algorithm}
\usepackage{algpseudocode}
\usepackage{amsmath}
\usepackage{amssymb}   
\usepackage{amsfonts}  
\usepackage{float}
\usepackage{booktabs}
\usepackage{multirow}

\title{EviTrack: Selection over Sampling for Delayed Disambiguation}

\author{%
   Omer A. Haq\thanks{Independent researcher.} \\
  Independent Researcher \\
  Baltimore, MD, USA \\
  \texttt{haqomer1@gmail.com}
}

\begin{document}

\maketitle



\begin{abstract}
Sequential prediction is challenging in regimes of \emph{delayed disambiguation},
where early observations are ambiguous and multiple latent explanations remain
plausible until sufficient evidence accumulates. Standard approaches based on
marginal inference struggle in this setting, either collapsing uncertainty
prematurely or failing to recover once informative evidence arrives.

We introduce \textbf{EviTrack}, a test-time inference framework that operates over
\emph{latent trajectories} rather than marginal states. EviTrack maintains a set of
competing trajectory hypotheses and applies evidence- and likelihood-ratio–based
selection to delay commitment until supported by data, drawing inspiration from
hypothesis management in multiple hypothesis tracking and track-before-detect.

To evaluate this setting, we construct a controlled synthetic benchmark with known
latent ground truth that explicitly exhibits delayed disambiguation. At matched
inference budget, EviTrack substantially outperforms sampling-based baselines,
achieving faster post-disambiguation recovery.

These results show that, in delayed disambiguation regimes, moderate trajectory-level
selection is more effective than increasing sampling coverage, highlighting selection
over sampling as a key principle for reliable sequential inference.
\end{abstract}

\section{Introduction}

Sequential prediction under nonstationary dynamics remains a central challenge in machine
learning. In many real-world settings, the data-generating process evolves over time,
producing ambiguous observations whose consequences only become apparent after sufficient
evidence has accumulated. Reliable prediction in these regimes requires inference procedures
that can maintain uncertainty without committing too early.

Most existing sequential models address uncertainty by marginalizing latent variables at
each timestep. Classical Bayesian filtering methods maintain a belief over the current
latent state conditioned on past observations~\cite{kalman1960new,doucet2000sequential},
while modern neural sequence models rely on amortized variational inference to approximate
per-timestep posteriors~\cite{chung2015recurrent,krishnan2017structured}. These approaches
share a common limitation: latent uncertainty is collapsed locally in time. When early
observations are weakly informative, competing explanations are averaged or discarded
prematurely, leading to brittle predictions and poor recovery when the underlying regime
shifts.

Many sequential problems are better characterized by \emph{delayed disambiguation}: multiple
latent trajectories are initially plausible and only become distinguishable after
accumulating evidence over extended horizons. This failure mode is well understood in
classical track-before-detect and hypothesis-tree methods in signal
processing~\cite{reed1974overview,barshalom2009tracking}, where maintaining explicit
hypotheses over time is essential for robust inference. Despite its importance,
trajectory-level inference has received limited attention in modern neural sequence modeling.

We introduce \textbf{EviTrack}, a general inference paradigm based on
\emph{trajectory-level evidence tracking}. Rather than approximating marginal filtering
distributions, EviTrack maintains a bounded set of candidate latent trajectory prefixes,
scores them using accumulated predictive evidence, and prunes implausible hypotheses in a
principled manner. This allows ambiguity to persist when warranted and collapse sharply when
the data becomes informative.

We evaluate EviTrack on a controlled synthetic benchmark designed to exhibit delayed
disambiguation in a tractable setting where ground-truth latent posteriors are available in
closed form. Rigorous evaluation of delayed disambiguation requires access to exact
posterior quantities and ground-truth disambiguation times that cannot be obtained from
real-world data, making synthetic benchmarks necessary to isolate the effects of interest.
Under matched inference budget, EviTrack substantially outperforms sequential importance
sampling and bootstrap particle filtering across forecasting, filtering, and hypothesis
quality metrics. Ablations show that the gains are robust across scoring functions,
branching factors, and budget allocations, and that EviTrack with reduced compute continues
to outperform particle filtering baselines at higher budgets.

\section{Related Work}

\paragraph{Sequential Bayesian filtering.}
Classical Bayesian filtering methods recursively update a posterior belief over the current
latent state. The Kalman filter and its extensions maintain parametric beliefs under linear
or locally linear assumptions~\cite{kalman1960new}, while particle filters approximate the
filtering distribution using weighted samples~\cite{doucet2000sequential,barshalom2009tracking}.
These methods are fundamentally designed to approximate the marginal posterior
$p(z_t \mid x_{1:t})$, so incompatible latent explanations are marginalized or resampled
away---leading to premature loss of multimodality under delayed disambiguation.

\paragraph{Track-before-detect and hypothesis-tree methods.}
In signal processing, track-before-detect (TBD) and multiple hypothesis tracking frameworks
explicitly maintain and score competing trajectories over
time~\cite{reed1974overview,barshalom2009tracking}, enabling robust inference under low
signal-to-noise and delayed detection. However, classical TBD approaches rely on
handcrafted dynamics and domain-specific likelihood models, and do not incorporate learned
representations or amortized proposals.

\paragraph{Search and inference in latent spaces.}
Search-based methods such as beam search and Monte Carlo Tree Search have been applied to
structured prediction and planning in learned latent spaces. These approaches optimize
reward or cost objectives and are not designed for probabilistic inference over latent
trajectories. EviTrack instead uses accumulated predictive evidence---rather than
reward---to guide hypothesis selection and pruning.

\paragraph{Positioning of EviTrack.}
EviTrack bridges classical hypothesis-tree inference and modern sequential latent-variable
models. Unlike Bayesian filtering, it does not marginalize latent uncertainty locally in
time. Unlike classical TBD methods, it operates under a learned generative model with
analytically tractable densities. Our contribution is a general inference paradigm for
sequential prediction: explicit evidence tracking over latent trajectory prefixes.

\begin{figure}[t]
    \centering
    \includegraphics[width=\linewidth]{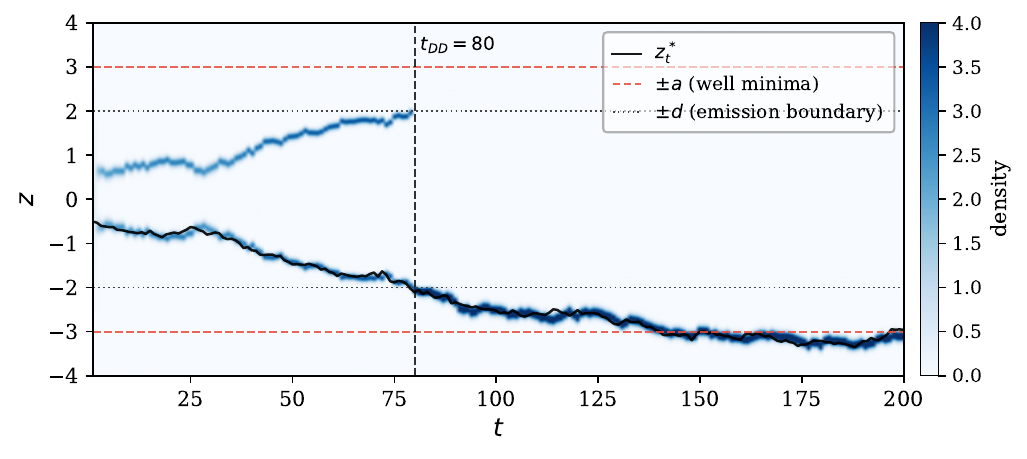}
    \caption{
    \textbf{Ground-truth filtering distribution exhibiting delayed disambiguation.}
    Heatmap shows the exact posterior $p(z_t \mid x_{1:t})$ computed via quadrature for a representative trajectory.
    The black curve denotes the true latent path $z_t^\ast$.
    The system exhibits two competing modes prior to the disambiguation time $t_{\mathrm{DD}}$ (vertical dashed line), corresponding to the two wells at $\pm a$ (red dashed lines).
    Observations become informative only when the latent crosses the emission boundary $\pm d$ (gray dotted lines), at which point the posterior collapses to a single mode.
    This delayed collapse highlights the failure mode of marginal filtering methods, which tend to commit prematurely before sufficient evidence accumulates.
    }
    \label{fig:dd_heatmap}
\end{figure}

\begin{figure}[t]
    \centering
    \includegraphics[width=\linewidth]{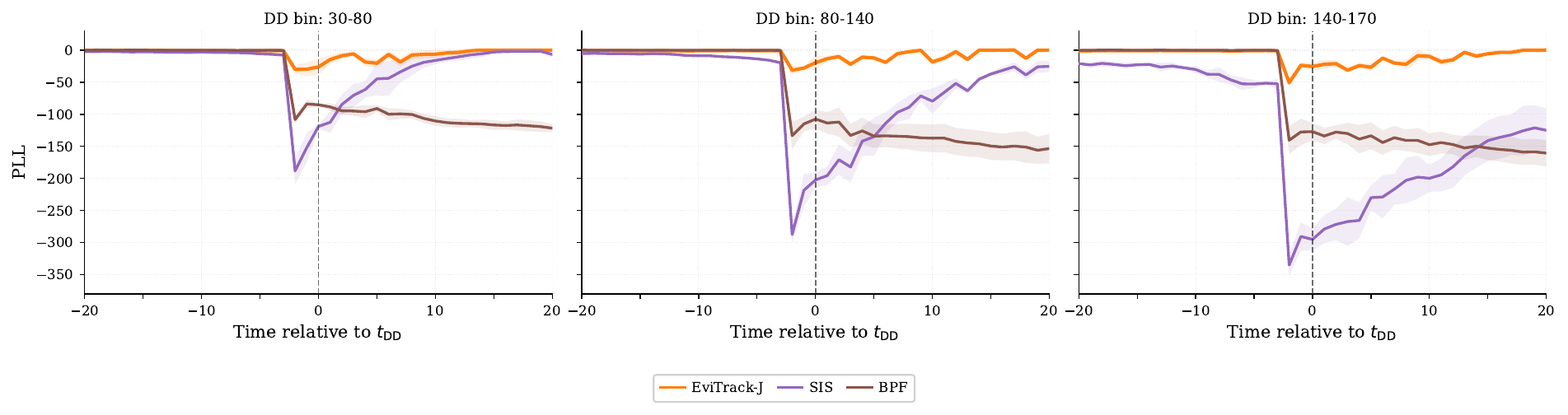}
    \caption{
    \textbf{Predictive log-likelihood (PLL) aligned to disambiguation time.}
    Mean PLL for one-step-ahead prediction ($H=1$) as a function of time relative to the true disambiguation time $t_{\mathrm{DD}}$, shown across different DD bins.
    For each seed, trajectories are aligned by $t_{\mathrm{DD}}$ and averaged within each bin over a window $t-t_{\mathrm{DD}}\in[-20,20]$; solid curves denote the mean across seeds and shaded regions denote one standard deviation.
    Prior to $t_{\mathrm{DD}}$, all methods perform similarly, reflecting inherent ambiguity in the observations.
    After disambiguation, EviTrack-J rapidly improves and approaches optimal performance, while SIS exhibits delayed recovery and Bootstrap PF (BPF) fails to recover.
    This demonstrates that maintaining multiple trajectory hypotheses enables faster and more accurate adaptation once informative evidence becomes available.
    }
    \label{fig:main_pll}
\end{figure}

\begin{figure}[t]
    \centering
    \includegraphics[width=\linewidth]{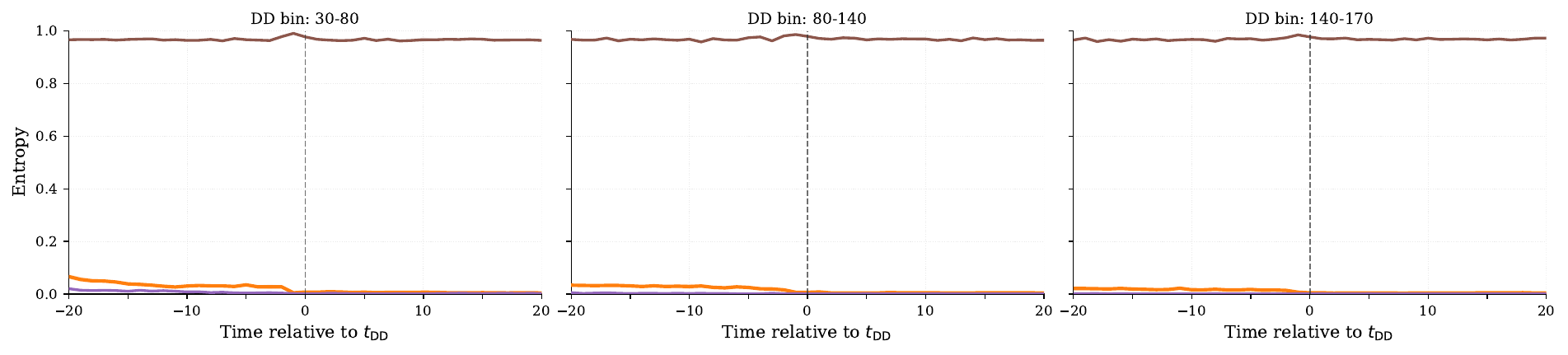}
    \vspace{-0.4em}
    \includegraphics[width=\linewidth]{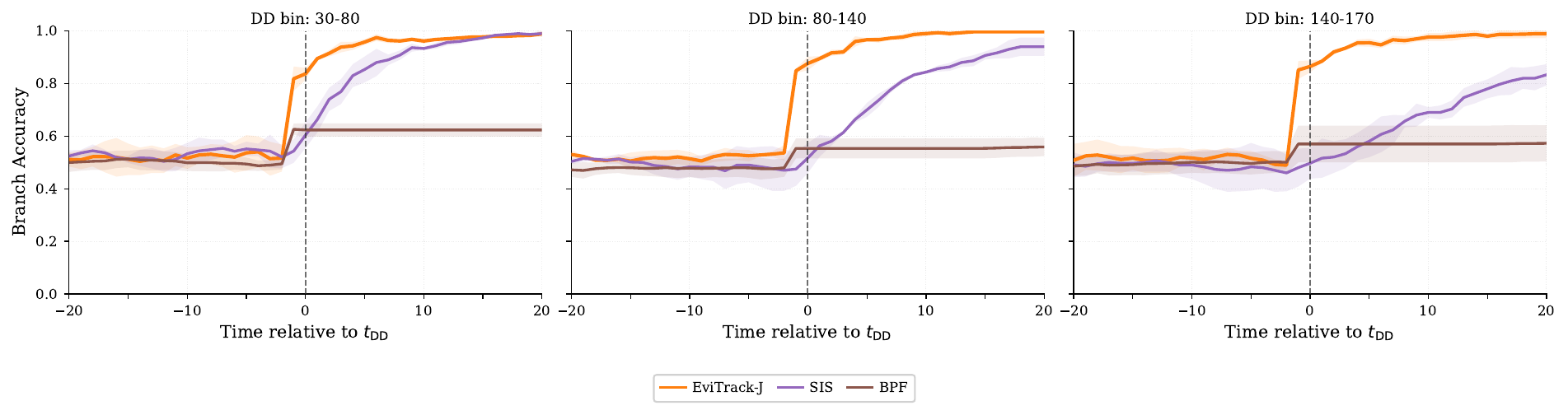}
    \caption{
    \textbf{Hypothesis uncertainty and filtering accuracy around disambiguation.}
    \textbf{Top:} normalized weight entropy.
    \textbf{Bottom:} filtering branch accuracy.
    Curves are aligned to the true disambiguation time $t_{\mathrm{DD}}$ and averaged within each DD bin as in Fig.~\ref{fig:main_pll}.
    Prior to $t_{\mathrm{DD}}$, Bootstrap PF maintains near-uniform weights, while EviTrack-J concentrates probability mass on a small set of coherent trajectories.
    At disambiguation, EviTrack rapidly commits to the correct hypothesis, producing a sharp increase in branch accuracy.
    SIS exhibits delayed recovery, while BPF fails to disambiguate effectively.
    }
    \label{fig:main_entropy_ba}
\end{figure}

\section{Problem Setup and Motivation}
\label{sec:problem}

We study \emph{test-time inference} for sequential latent-variable models in regimes with
\emph{delayed disambiguation}: early observations admit multiple incompatible latent explanations
that only become separable after sufficient evidence accumulates.

\subsection{Sequential latent-variable models}

We consider a sequential latent-variable model over observations $x_{1:T}$
and latent variables $z_{1:T}$ with transition $p(z_t \mid z_{<t})$
and emission $p(x_t \mid x_{<t}, z_{\le t})$.
The full causal factorization of the joint distribution is provided
in Appendix~\ref{app:wm_factorization}.

\subsection{Motivation: why marginal state inference is brittle}

Standard filtering targets the marginal belief $p(z_t\mid x_{1:t})$, whereas prediction
under a latent world model is a mixture over trajectory prefixes:
\begin{equation}
p(x_{t+1}\mid x_{1:t})
=
\int p(x_{t+1}\mid z_{1:t},x_{1:t})\,p(z_{1:t}\mid x_{1:t})\,dz_{1:t}.
\end{equation}

When early observations are ambiguous, $p(z_{1:t}\mid x_{1:t})$ is multi-modal.
Particle filters can represent this multimodality, but both standard variants exhibit
failure modes under delayed disambiguation: resampling (BPF) can eliminate valid
trajectories prematurely, while non-resampling methods (SIS) suffer from weight
degeneracy and lack a mechanism to refine competing hypotheses. In both cases,
the method fails to maintain and improve plausible trajectories over time.

\paragraph{EviTrack viewpoint.}
EviTrack maintains explicit trajectory hypotheses and applies moderate selection,
preserving ambiguity early while refining hypotheses as evidence accumulates.

\subsection{Trajectory scoring}

Each trajectory $z_{1:t}$ is assigned a scalar score used for both hypothesis
selection and predictive weighting. We consider three variants:
\begin{align}
J_t(z_{1:t}; x_{1:t}) 
&= \log p(x_{1:t}, z_{1:t}), \\
E_t(z_{1:t}; x_{1:t}) 
&= \log p(x_{1:t}\mid z_{1:t}), \\
J_t^{\mathrm{tbd}}(z_{1:t}; x_{1:t}) 
&= \log p(x_{1:t}\mid z_{1:t}) 
   + \log \frac{p(z_{1:t})}{p_0(z_{1:t})},
\end{align}
where $p_0(z_{1:t})$ is a Gaussian random-walk prior,
$p_0(z_1)=\mathcal{N}(0,\sigma_{\mathrm{bg}}^2 I)$ and
$p_0(z_t\mid z_{t-1})=\mathcal{N}(z_{t-1},\sigma_{\mathrm{bg}}^2 I)$.

All scores accumulate additively over time and can be interpreted as
log-likelihood ratios under different null models. In practice, these scores
induce similar qualitative behavior, with differences primarily affecting
trajectory ranking after disambiguation (see Appendix~\ref{app:scoring} and
Sec.~\ref{sec:ablations}).

\section{Method: EviTrack}
\label{sec:method}

EviTrack is a general test-time inference procedure for sequential latent-variable models.
It maintains a finite hypothesis set of latent trajectory prefixes, expands them using a
proposal distribution, accumulates predictive evidence, prunes to a fixed budget, and
predicts by a trajectory mixture.

\subsection{Hypothesis sets}

At time $t$, EviTrack maintains a hypothesis set of $K$ retained trajectory prefixes
\begin{equation}
\mathcal H_t \;=\; \{ z^{(i)}_{1:t} \}_{i=1}^K,
\label{eq:Ht_main}
\end{equation}
together with associated (unnormalized) scores
\begin{equation}
S_t^{(i)} := S_t(z^{(i)}_{1:t}; x_{1:t}).
\end{equation}
We consider several choices of trajectory score, including the evidence score,
joint score, and likelihood-ratio variants:
\begin{equation}
S_t^{(i)} \in \Big\{
E_t(z^{(i)}_{1:t}; x_{1:t}),\;
J_t(z^{(i)}_{1:t}; x_{1:t}),\;
J_t^{\mathrm{tbd}}(z^{(i)}_{1:t}; x_{1:t})
\Big\}.
\label{eq:score_choice_main}
\end{equation}

\subsection{Hypothesis expansion and score accumulation}

Given $\mathcal H_t$, each retained prefix is branched into $C$ candidate continuations by
sampling from the latent transition model:
\begin{equation}
z^{(i,j)}_{t+1} \sim p(\cdot \mid z^{(i)}_{t}),
\qquad j=1,\dots,C.
\label{eq:branch_main}
\end{equation}
For learned world models, one may instead use an amortized proposal
$q_\phi(\cdot \mid z^{(i)}_{1:t}, x_{1:t})$ to concentrate candidates in regions favored by the posterior.

Each child prefix $z^{(i,j)}_{1:t+1} = (z^{(i)}_{1:t}, z^{(i,j)}_{t+1})$
inherits its parent score and receives an incremental contribution:
\begin{equation}
\Delta_{t+1}^{(i,j)}
:=
S_{t+1}(z^{(i,j)}_{1:t+1}; x_{1:t+1})
-
S_t(z^{(i)}_{1:t}; x_{1:t}),
\label{eq:delta_main}
\end{equation}
so that scores update additively:
\begin{equation}
S^{(i,j)}_{t+1}
=
S^{(i)}_{t} + \Delta^{(i,j)}_{t+1}.
\label{eq:additive_score_main}
\end{equation}

The candidate pool at time $t+1$ consists of all $KC$ children,
denoted $\widetilde{\mathcal H}_{t+1}$.

\subsection{Local and global pruning}

To control the exponential growth of trajectory hypotheses, EviTrack employs two forms of pruning.

\paragraph{Local pruning (selection).}
During expansion, each parent trajectory $z^{(i)}_{1:t}$ generates $C$ candidate children
by sampling from the transition model,
\(
z^{(i,j)}_{t+1} \sim p(z_{t+1} \mid z^{(i)}_{1:t})
\).
Rather than retaining all $C$ continuations, EviTrack selects a single child per parent
according to the scoring function:
\begin{equation}
z^{(i,\star)}_{1:t+1}
=
\arg\max_{j \in \{1,\dots,C\}}
S_{t+1}(z^{(i,j)}_{1:t+1}; x_{1:t+1}).
\end{equation}
This reduces the candidate set from $KC$ to $K$ trajectories while preserving one continuation per parent.

This operation induces an order-statistics–biased distribution over transition samples.
Let $F(s)$ denote the cumulative distribution function of the induced score
$S(z_{t+1})$ under the transition $p(z_{t+1} \mid z_{1:t})$. Then the selected child
is distributed according to
\begin{equation}
p_C^\star(z_{t+1} \mid z_{1:t})
=
C\,p(z_{t+1} \mid z_{1:t})\,F(S(z_{t+1}))^{C-1},
\end{equation}
which increasingly concentrates on high-scoring continuations as $C$ grows
(see Appendix~\ref{app:order_stats} for derivation). Thus, local pruning
interpolates between stochastic sampling ($C=1$) and near-deterministic MAP
selection as $C \to \infty$, formalizing the trade-off between exploration
and greedy selection.

\paragraph{Global pruning.}
In addition to local selection, EviTrack may apply global pruning across trajectories.
At selected time steps, the hypothesis set is updated by retaining only the top-$K$ trajectories:
\begin{equation}
\mathcal H_{t}
\leftarrow
\textsc{TopK}(\mathcal H_{t}, K).
\end{equation}
This operation enforces a global budget by removing low-scoring trajectories, but may
discard valid hypotheses prematurely if applied too aggressively.

\paragraph{Discussion.}
Local pruning operates within each trajectory branch, while global pruning acts across
the entire hypothesis set. In the limit $C=1$ and no global pruning, the method
reduces to SIS.

\subsection{Prediction and generality}

EviTrack forms predictions as a weighted mixture over the retained trajectory hypotheses,
where weights $\pi_t^{(i)} \propto \exp(S_t^{(i)})$ correspond to a normalized,
support-restricted approximation of the posterior over trajectory prefixes. Future
predictions are computed via Monte Carlo rollouts from the transition prior. In our
experiments, all quantities are evaluated under the true model; for learned world models,
the same inference procedure applies with a learned transition and optional proposal.

\section{Experiments}
\label{sec:experiments}

We evaluate EviTrack on a controlled synthetic benchmark designed to exhibit delayed
disambiguation in a tractable setting where ground-truth dynamics are known exactly.
All methods operate under the same fixed model and differ only in how latent uncertainty
is represented and propagated at test time.

\subsection{Double-Well Delayed-Disambiguation Benchmark}
\label{sec:synthetic}

\paragraph{Motivation.}
The benchmark is designed to isolate the core phenomenon targeted by EviTrack:
\emph{delayed disambiguation}.
Early observations are consistent with multiple latent explanations, none of which can be
ruled out locally.
Only after sufficient evidence has accumulated do these explanations become distinguishable.
This setting directly exposes the failure mode of marginal filtering methods, which tend to
collapse uncertainty prematurely and commit to an explanation before it is justified.

\paragraph{Latent dynamics.}
The latent trajectory evolves according to
\begin{equation}
z_t \sim p(z_{t}|z_{t-1})=\mathcal{N}\!\big(\mu(z_{t-1}),\;\sigma_z^2\big),
\label{eq:doublewell_transition}
\end{equation}
with drift
\begin{equation}
\mu(z) = z - \Delta t\,V_0\, z(z^2 - a^2),
\end{equation}
corresponding to a symmetric double-well system with stable equilibria at $z=\pm a$.
Trajectories initialized near $z=0$ evolve under competing basin attractions before
committing to one well.

\paragraph{Emission model.}
Observations are generated as
\begin{equation}
x_t \sim p(x_t|z_t)= \mathcal{N}\!\big(h(z_t),\;\sigma_x^2\big), \qquad
h(z)=
\begin{cases}
z^2, & |z|\le d,\\
z,   & |z|>d.
\end{cases}
\end{equation}
Within the region $|z| \le d$, the map is even: $h(z) = h(-z)$, so trajectories in
opposite wells produce statistically identical observations.
When the well minima satisfy $a > d$, trajectories initialized near $z = 0$ are
attracted to $\pm a$ and remain in the non-injective region, making the two basin
hypotheses observationally indistinguishable for an extended period.
Disambiguation only becomes possible once the latent fluctuates beyond $|z| > d$,
where $h$ is injective and sign information is recoverable.
The condition $a > d$ is therefore what produces delayed disambiguation: it ensures
the wells lie outside the ambiguous region, so that evidence accumulates slowly and
collapse is genuinely deferred rather than immediate.

\paragraph{Analytical world model.}
Both transition and emission densities are Gaussian and available in closed form, so
inference is performed under the exact ground-truth model with no approximation of $p$.
Hypothesis expansion uses the transition prior \eqref{eq:doublewell_transition} as the
proposal throughout.

\paragraph{Delayed disambiguation detection.}
Figure~\ref{fig:dd_heatmap} shows the exact posterior $p(z_t \mid x_{1:t})$ computed via
quadrature for a representative trajectory drawn from the world model, illustrating the
phenomenon we seek to detect: the posterior remains bimodal for an extended period before
collapsing sharply to the correct basin at $t_{\mathrm{DD}}$. We define the disambiguation
time $t_{\mathrm{DD}}$ as the first timestep at which the posterior probability assigned to
the correct basin exceeds a threshold $\tau$. We exclude trajectories with near-immediate
($t_{\mathrm{DD}} < 30$) or near-terminal ($t_{\mathrm{DD}} > 170$) disambiguation and
partition the remainder into three regimes: \textbf{Early} ($30 \le t_{\mathrm{DD}} < 80$),
\textbf{Mid} ($80 \le t_{\mathrm{DD}} < 140$), and \textbf{Late}
($140 \le t_{\mathrm{DD}} \le 170$). A balanced evaluation set of 100 trajectories per bin
is used.

\paragraph{Experimental setup.}
We compare EviTrack-J against particle filtering baselines under matched inference
budget $N = K \cdot C = 64$. EviTrack-J uses $K=32$ trajectories with branching factor
$C=2$, while SIS and BPF use $N=64$ particles, isolating selection among candidate
continuations versus independent sampling. All other experimental parameters are fixed
and given in Table~\ref{tab:params}.

\textbf{SIS} propagates particles from the transition prior with sequential importance
weighting and no resampling, corresponding to EviTrack-E with $C=1$ (no selection).
\textbf{BPF} additionally performs ESS-triggered resampling, which removes low-weight
trajectories during inference. All methods accumulate evidence identically but differ
in hypothesis management.

Models condition on $x_{1:t}$ and predict $x_{t+1:t+H}$ in a rolling forecasting setting,
with $H \in \{1,5,10\}$. For each inference seed and delayed-disambiguation (DD) regime,
trajectories are aligned to $t_{\mathrm{DD}}$ (20 steps before and after), averaged across
trajectories, and then aggregated across seeds; shaded regions denote standard deviation
across seeds.

We report predictive log-likelihood (PLL, $H=1$), filtering branch accuracy (BA), and
weight entropy; full metric definitions are provided in Appendix~\ref{app:metrics}.
Additional metrics (including $H \in \{5,10\}$ horizons and MSE) exhibit consistent
trends and are omitted for clarity.

\paragraph{Main results.}
The main comparison is shown in Figures.~\ref{fig:main_pll}--\ref{fig:main_entropy_ba}. 
Before the disambiguation time $t_{\mathrm{DD}}$, all methods operate near the intrinsic 
ambiguity limit: observations are compatible with both latent basins, and filtering branch 
accuracy remains close to chance. After $t_{\mathrm{DD}}$, however, the methods separate 
sharply. EviTrack-J rapidly improves in PLL and commits to the 
correct basin, while SIS recovers more slowly and BPF largely fails to disambiguate. 
This behavior is quantified in Tables~\ref{tab:main_ba_all} and~\ref{tab:main_pll_all}: 
post-DD filtering branch accuracy reaches $0.987$ for EviTrack-J, compared with $0.912$ 
for SIS and $0.584$ for BPF, while post-DD PLL is $-2.948$ for EviTrack-J, compared with 
$-54.254$ for SIS and $-154.155$ for BPF.

These results support the central claim that, at fixed compute, evidence-guided selection 
is more effective than increasing sampling coverage alone. SIS is the closest baseline to 
EviTrack-J because it propagates full trajectories without resampling, thereby preserving 
trajectory identity and allowing evidence to accumulate over time. In contrast, BPF uses ESS-triggered resampling: when the effective sample size falls below the threshold, low-weight trajectories are resampled away. In delayed-disambiguation regimes, this can discard valid hypotheses before later evidence becomes informative, leading to poor post-DD recovery.

EviTrack-J builds on the same trajectory-level evidence accumulation as SIS but augments it 
with local selection. With $K=32$ retained trajectories and branching factor $C=2$, each 
hypothesis evaluates multiple candidate continuations and selects the highest-scoring child, 
improving trajectory quality without sacrificing diversity. This additional selection step 
yields a sharper post-DD transition in both predictive likelihood and branch accuracy, 
demonstrating that selection on top of trajectory preservation is more effective than 
sampling alone.

\section{Ablations}
\label{sec:ablations}

EviTrack defines a family of inference procedures parameterized by a small number of
design choices: the scoring function used for pruning and weighting, the global pruning
interval, and the allocation of the inference budget between hypothesis count and branching
factor. All variants operate under the same fixed world model and differ only in test-time
inference.

\subsection{Scoring functions}

We ablate the trajectory scoring rule used for both pruning and predictive weighting by
comparing the three variants introduced in Sec.~\ref{sec:problem}: evidence-only
(EviTrack-E), joint (EviTrack-J), and background-normalized (EviTrack-TBD) at fixed $K=32, C=2,$ and $G=\infty$. In all cases,
the same score governs hypothesis selection and mixture weighting, with all other inference
components held fixed.

Appendix~\ref{app:ablation:scoring} Figure.~\ref{fig:scoring_ablation_ba}--\ref{fig:scoring_ablation_pll} shows that all scoring variants exhibit the same
qualitative behavior around $t_{\mathrm{DD}}$: performance is similar before disambiguation
and improves sharply once informative evidence arrives. The quantitative picture is provided
in Appendix~\ref{app:tables:scoring}.
Tables~\ref{tab:scoring_ba_all}--\ref{tab:scoring_ba_bins} report filtering branch accuracy
before and after disambiguation, aggregated across all DD bins and broken out per bin
respectively; Tables~\ref{tab:scoring_pll_all}--\ref{tab:scoring_pll_bins} give the
corresponding PLL results. Across all bins and both metrics, the differences between scoring
rules are small relative to the gap between EviTrack and the particle-filtering baselines.
Evidence-only scoring achieves the best aggregate post-DD PLL
(Table~\ref{tab:scoring_pll_all}: $-2.459$ for EviTrack-E vs.\ $-2.948$ for EviTrack-J),
while all three variants reach near-identical post-DD filtering branch accuracy
(Table~\ref{tab:scoring_ba_all}: $0.987$--$0.990$). The main effect is therefore not the
particular score choice, but the use of trajectory-level selection itself.

\subsection{Global pruning interval}
\label{sec:abl_global_pruning}

We ablate the frequency of global top-$K$ pruning by varying the interval
\begin{equation}
G \in \{1,\, 5,\, 10,\, 20,\, \infty\},
\end{equation}
with $K = 32$ and $C = 2$ fixed. Global pruning is applied every $G$ steps; at intermediate
steps each hypothesis retains only its highest-scoring child (local pruning). The case $G=1$
corresponds to pruning at every step, while $G=\infty$ disables global competition entirely.
We evaluate EviTrack-J for all settings.

Appendix~\ref{app:ablation:g}
Figures.~\ref{fig:app_g_ablation_ba}--\ref{fig:app_g_ablation_pll} shows the filtering BA and PLL ($H=1$), respectively. Performance improves monotonically as $G$ increases. Frequent global
pruning ($G = 1, 5$) leads to poor post-DD recovery: cross-hypothesis competition eliminates
valid hypotheses before later evidence becomes informative. Larger values allow hypotheses to
persist and accumulate evidence before competing globally. The best performance is achieved
at $G = \infty$, indicating that premature cross-hypothesis elimination---not particle count
or scoring choice---is the primary failure mode in delayed disambiguation settings. This also
motivates the use of $G = \infty$ for EviTrack-J in the main experiment.

\subsection{Branching factor}
\label{sec:abl_children}

Fixing the total budget $N = K \times C = 64$ and $G = 1$, we vary how the budget is split
between hypothesis count $K$ and branching factor $C$:
\begin{equation}
C \in \{2,\, 4,\, 8,\, 16,\, 32\},
\end{equation}
with $K$ adjusted accordingly. Large $K$, small $C$ favors breadth---maintaining many
diverse hypotheses with shallow per-step selection; small $K$, large $C$ favors
depth---fewer hypotheses each explored more aggressively at each step. We consider only EviTrack-J for each sweep.

Appendix~\ref{app:ablation:c} Figures.~\ref{fig:c_ablation_ba}--\ref{fig:c_ablation_pll} shows BA and PLL ($H=1$) aligned to $t_{\mathrm{DD}}$ across all three DD
bins. Small branching factors ($C = 2, 4$) achieve the best post-DD PLL performance, while
larger values ($C \geq 16$) degrade sharply. The complementary filtering branch accuracy tells the same story: excessive local selection reduces hypothesis diversity prior to
disambiguation and impairs recovery once informative evidence arrives. This demonstrates a
tradeoff---moderate selection improves trajectory quality by exploiting score variability at
each step, but aggressive selection collapses the hypothesis set toward a single trajectory
before the ambiguity is genuinely resolved.


\subsection{Inference budget scaling}
\label{sec:abl_budget}

We fix the branching factor $C = 2$ and sweep hypothesis count
\begin{equation}
K \in \{2,\, 4,\, 8,\, 16,\, 32,\, 64\},
\end{equation}
with $G = \infty$ fixed, assessing how EviTrack performance scales with total inference budget
$N = K \times C$. At each budget level we include matched BPF and SIS baselines with
$N = K \times C$ particles, isolating the contribution of evidence-guided hypothesis
management from raw particle count.

The results are shown in Appendix~\ref{app:ablation:k}
Figures.~\ref{fig:app_k_ablation_ba}--\ref{fig:app_k_ablation_pll} shows filtering BA and PLL, respectively. EviTrack improves
rapidly with moderate $K$, achieving strong post-DD performance at relatively small budgets.
In contrast, both BPF and SIS exhibit substantially lower filtering accuracy and PLL across all budgets and regimes. Increasing particle count in SIS and BPF does
not close the performance gap with EviTrack, even at higher budgets. This demonstrates that
improved sampling coverage alone is insufficient and that structured hypothesis
selection---not particle count---is the dominant factor.

\section{Discussion and Conclusion}
\label{sec:conclusion}

The central finding of this work is that \emph{trajectory-level selection is more effective
than increasing sampling coverage} in regimes with delayed disambiguation. In the main
comparison (Figures~\ref{fig:main_pll}--\ref{fig:main_entropy_ba}), all methods operate
under matched compute $N = K \cdot C = 64$. EviTrack-J ($K=32$, $C=2$, $G=\infty$)
substantially outperforms SIS and BPF ($N=64$), achieving post-DD PLL of $-2.9$
(vs.\ $-54$, $-154$) and branch accuracy of $0.987$ (vs.\ $0.912$, $0.584$)
(Tables~\ref{tab:main_pll_all},~\ref{tab:main_ba_all}).

This gap arises from how hypotheses are managed. EviTrack preserves trajectory identity
and defers competition, allowing valid hypotheses to survive until disambiguating evidence
arrives. In contrast, ESS-triggered resampling in BPF can eliminate correct trajectories
prematurely, while SIS suffers from weight degeneracy. Ablations confirm this behavior:
performance improves with minimal pruning ($G=\infty$) and small branching factors
($C=2,4$), while increasing particle count in SIS and BPF does not close the gap
(Figures~\ref{fig:app_g_ablation_ba}--\ref{fig:app_k_ablation_pll}). 

Scoring variants exhibit consistent qualitative behavior, with modest differences across
joint, evidence-only, and background-normalized objectives (Appendix~\ref{app:ablation:scoring}, \ref{app:tables:scoring});
while certain variants improve predictive likelihood in some regimes, these effects are
small relative to the gap between EviTrack and particle filtering baselines, indicating
that hypothesis management is the dominant factor. Together, these results show that
\emph{how compute is allocated matters more than how much compute is used} when observations
are ambiguous.

\paragraph{Limitations.}
Performance depends on $K$, $C$, pruning schedule, and scoring; adaptive selection remains
open. EviTrack is not a consistent Monte Carlo estimator, as pruning approximates posterior
support rather than integrating over it. Evaluation relies on synthetic settings with
latent ground truth; extending to learned world models and high-dimensional observations
is an important direction.

\paragraph{Conclusion.}
EviTrack performs inference over latent trajectories, combining evidence accumulation with
moderate selection to delay commitment until sufficient information is available. This
yields faster and more reliable post-disambiguation recovery than sampling-based methods,
highlighting selection over sampling as a key principle for sequential inference under
delayed disambiguation.



{\small
\bibliographystyle{unsrt}
\bibliography{references}

@article{kalman1960new,
  title={A new approach to linear filtering and prediction problems},
  author={Kalman, Rudolf E},
  journal={Journal of Basic Engineering},
  volume={82},
  number={1},
  pages={35--45},
  year={1960}
}

@article{doucet2000sequential,
  title={Sequential Monte Carlo methods in practice},
  author={Doucet, Arnaud and Godsill, Simon and Andrieu, Christophe},
  journal={Statistics and Computing},
  volume={10},
  number={3},
  pages={197--208},
  year={2000}
}

@inproceedings{chung2015recurrent,
  title={A recurrent latent variable model for sequential data},
  author={Chung, Junyoung and Kastner, Kyle and Dinh, Laurent and Goel, Kratarth and Courville, Aaron},
  booktitle={Advances in Neural Information Processing Systems},
  year={2015}
}

@inproceedings{krishnan2017structured,
  title={Structured inference networks for nonlinear state space models},
  author={Krishnan, Rahul and Shalit, Uri and Sontag, David},
  booktitle={Proceedings of the AAAI Conference on Artificial Intelligence},
  year={2017}
}

@book{barshalom2009tracking,
  title={Tracking and Data Fusion},
  author={Bar-Shalom, Yaakov and Willett, Peter and Tian, Xin},
  publisher={YBS Publishing},
  year={2009}
}

@article{reed1974overview,
  title={An Overview of Track-Before-Detect Techniques},
  author={Reed, I. S. and Gagliardi, R. M.},
  journal={IEEE Transactions on Aerospace and Electronic Systems},
  year={1974}
}
}


\newpage

\appendix

\section{World Model Factorization}
\label{app:wm_factorization}

Let $x_{1:T}$ denote observations and $z_{1:T}$ latent variables.
We begin from the generic identity
\begin{equation}
p(x_{1:T}, z_{1:T}) \;=\; p(z_{1:T})\,p(x_{1:T}\mid z_{1:T}).
\label{eq:joint_basic}
\end{equation}

Applying the chain rule to the latent trajectory imposes no Markov restriction:
\begin{equation}
p(z_{1:T}) \;=\; p(z_1)\prod_{t=2}^T p(z_t \mid z_{<t}),
\label{eq:prior_chain_main}
\end{equation}

and similarly for the conditional likelihood,
\begin{equation}
p(x_{1:T}\mid z_{1:T})
\;=\;
\prod_{t=1}^T p(x_t \mid x_{<t}, z_{1:T}).
\label{eq:like_chain_main}
\end{equation}

To obtain a \emph{causal} generative process, we restrict emissions to be independent of future latents,
\begin{equation}
p(x_t \mid x_{<t}, z_{1:T})
\;=\;
p(x_t \mid x_{<t}, z_{\le t}),
\label{eq:causal_emission_main}
\end{equation}

which yields the causal joint factorization
\begin{equation}
p(x_{1:T}, z_{1:T})
=
\Big[p(z_1)\prod_{t=2}^T p(z_t \mid z_{<t})\Big]
\cdot
\Big[\prod_{t=1}^T p(x_t \mid x_{<t}, z_{\le t})\Big].
\label{eq:causal_joint_main}
\end{equation}

\section{Trajectory scoring functions}
\label{app:scoring}

We provide detailed definitions and interpretations of the trajectory scoring
functions used in EviTrack.

\paragraph{Joint score.}
The joint trajectory score is defined as
\begin{equation}
J_t(z_{1:t}; x_{1:t})
:= \log p(x_{1:t}, z_{1:t})
= \log p(z_{1:t}) + \log p(x_{1:t} \mid z_{1:t}).
\end{equation}
Under standard factorization,
\begin{align}
\log p(z_{1:t})
&= \log p(z_1) + \sum_{k=2}^t \log p(z_k \mid z_{<k}), \\
\log p(x_{1:t} \mid z_{1:t})
&= \sum_{k=1}^t \log p(x_k \mid x_{<k}, z_{\le k}).
\end{align}
This score balances data fit and prior plausibility.

\paragraph{Likelihood-ratio interpretation.}
More generally, trajectory scores can be written as log-likelihood ratios
\begin{equation}
\log \frac{p(x_{1:t}, z_{1:t})}{p_0(x_{1:t}, z_{1:t})},
\end{equation}
for a chosen null model $p_0$. Different choices of $p_0$ yield different
scoring rules.

\paragraph{Evidence score.}
Let the null model preserve latent dynamics but remove dependence on observations:
\begin{equation}
p_0^{\mathrm{ev}}(z_{1:t}) = p(z_{1:t}),
\qquad
p_0^{\mathrm{ev}}(x_{1:t} \mid z_{1:t}) = p_0(x_{1:t}).
\end{equation}
Then
\begin{equation}
\log \frac{p(x_{1:t}, z_{1:t})}{p_0^{\mathrm{ev}}(x_{1:t}, z_{1:t})}
= \log p(x_{1:t} \mid z_{1:t}) - \log p_0(x_{1:t}),
\end{equation}
which differs from
\begin{equation}
E_t(z_{1:t}; x_{1:t}) := \log p(x_{1:t} \mid z_{1:t})
\end{equation}
only by a constant independent of $z_{1:t}$.

\paragraph{Background-normalized score.}
We define a null model with a structureless latent process:
\begin{equation}
p_0(z_1) = \mathcal{N}(0, \sigma_{\mathrm{bg}}^2 I),
\qquad
p_0(z_t \mid z_{t-1}) = \mathcal{N}(z_{t-1}, \sigma_{\mathrm{bg}}^2 I),
\end{equation}
and independent observations. The resulting score is
\begin{equation}
J_t^{\mathrm{tbd}}(z_{1:t}; x_{1:t})
:= J_t(z_{1:t}; x_{1:t}) - \log p_0(z_{1:t}),
\end{equation}
which can be written as
\begin{equation}
J_t^{\mathrm{tbd}}
= \log p(x_{1:t}\mid z_{1:t})
+ \log \frac{p(z_{1:t})}{p_0(z_{1:t})}.
\end{equation}

\paragraph{Interpretation.}
The scoring rules correspond to progressively stronger normalization:
\begin{itemize}
\item $J_t$: joint likelihood under the model,
\item $E_t$: explanatory power relative to a latent-agnostic baseline,
\item $J_t^{\mathrm{tbd}}$: explanatory power relative to a structureless latent process.
\end{itemize}
All three scores induce similar qualitative behavior in EviTrack, with differences
primarily affecting trajectory ranking once disambiguating evidence becomes
available.

\section{Order-Statistics Interpretation of Local Pruning}
\label{app:order_stats}

We derive the distribution induced by local pruning when selecting the
highest-scoring child among $C$ transition samples.

Fix a parent trajectory $z_{1:t}$ and draw $C$ candidate children
\begin{equation}
z_1, \dots, z_C \overset{iid}{\sim} p(z_{t+1} \mid z_{1:t}).
\end{equation}
Define the score of each candidate as
\begin{equation}
S_j = S(z_j).
\end{equation}
Local pruning selects
\begin{equation}
z^\star = z_{J^\star}, \quad J^\star = \arg\max_{j \le C} S_j.
\end{equation}

We derive the distribution of $z^\star$. For a small region $dz$ around $z$,
\begin{equation}
\mathbb{P}(z^\star \in dz)
=
\sum_{j=1}^C \mathbb{P}(z_j \in dz, \; j = \arg\max).
\end{equation}
By symmetry,
\begin{equation}
=
C \, \mathbb{P}(z_1 \in dz, \; S_1 \ge S_2, \dots, S_C).
\end{equation}

Conditioning on $z_1 = z$ gives
\begin{equation}
\mathbb{P}(z^\star \in dz)
=
C \, p(z_{t+1}=z \mid z_{1:t})\,dz \cdot
\mathbb{P}(S_2 \le S(z), \dots, S_C \le S(z)).
\end{equation}
Since the remaining samples are independent,
\begin{equation}
\mathbb{P}(S_2 \le S(z), \dots, S_C \le S(z))
=
F(S(z))^{C-1},
\end{equation}
where
\begin{equation}
F(s) = \mathbb{P}_{z_{t+1} \sim p(\cdot \mid z_{1:t})}(S(z_{t+1}) \le s).
\end{equation}

Thus,
\begin{equation}
\mathbb{P}(z^\star \in dz)
=
C\,p(z_{t+1}=z \mid z_{1:t})\,F(S(z))^{C-1}\,dz,
\end{equation}
and the resulting density is
\begin{equation}
p_C^\star(z_{t+1} \mid z_{1:t})
=
C\,p(z_{t+1} \mid z_{1:t})\,F(S(z_{t+1}))^{C-1}.
\end{equation}

This shows that local pruning transforms the transition distribution into an
order-statistics–biased distribution that increasingly concentrates on
high-scoring regions as $C$ grows.

In the limit $C \to \infty$, assuming the transition has support near the maximizer,
\begin{equation}
z^\star \to \arg\max_z S(z),
\end{equation}
recovering deterministic MAP selection. For small $C$, the distribution remains
close to the transition, preserving stochastic exploration.

\section{Metric Definitions}
\label{app:metrics}

At each timestep $t$, the inference state is a hypothesis set
$\mathcal{H}_t = \{z^{(i)}_{1:t}\}_{i=1}^K$ with normalized mixture weights
$\pi_t^{(i)} = \exp(S_t^{(i)}) / \sum_j \exp(S_t^{(j)})$.
The filtering distribution over the current latent is approximated by this weighted
hypothesis set,
\begin{equation}
p(z_t \mid x_{1:t}) \;\approx\; \sum_{i=1}^K \pi_t^{(i)}\, \delta(z_t - z^{(i)}_t).
\label{eq:filt_approx}
\end{equation}
Forecasting metrics are computed for each prediction horizon
$H \in \{1, 5, 10\}$ via $M$ Monte Carlo rollouts from the transition prior.
Filtering metrics are computed directly from the hypothesis set at time $t$ without
rollouts.
All metrics are reported as functions of $t$ and aggregated within each DD regime.

\subsection{Forecasting metrics}

\paragraph{Predictive log-likelihood (PLL).}
The predictive log-likelihood at horizon $H$ is
\begin{equation}
\mathrm{PLL}_H = \log p(x_{t+H} \mid x_{1:t}),
\end{equation}
where the predictive density is obtained by marginalizing over the trajectory
posterior,
\begin{equation}
p(x_{t+H} \mid x_{1:t})
= \int p(x_{t+H} \mid z_{1:t+H},\, x_{1:t})\,
p(z_{t+1:t+H} \mid z_{1:t})\,
p(z_{1:t} \mid x_{1:t})\,
dz_{1:t+H}.
\end{equation}
Substituting the filtering approximation \eqref{eq:filt_approx} gives the mixture
form
\begin{equation}
p(x_{t+H} \mid x_{1:t})
\approx \sum_{i=1}^K \pi_t^{(i)}
\int p(x_{t+H} \mid z^{(i)}_{1:t+H},\, x_{1:t})\,
p(z_{t+1:t+H} \mid z^{(i)}_{1:t})\,
dz_{t+1:t+H},
\end{equation}
where the inner integral marginalizes over future latent continuations conditioned
on the retained prefix $z^{(i)}_{1:t}$.

\textit{Numerical evaluation.}
The inner integral is approximated by $M$ Monte Carlo rollouts from the transition
prior, $z^{(i,m)}_{t+1:t+H} \sim p(\cdot \mid z^{(i)}_{1:t})$, giving the
per-hypothesis log-likelihood estimate
\begin{equation}
\ell^{(i)}_H
= \log \frac{1}{M} \sum_{m=1}^M
p(x_{t+H} \mid z^{(i,m)}_{1:t+H},\, x_{1:t}).
\end{equation}
The mixture PLL is then computed in log-space for numerical stability,
\begin{equation}
\mathrm{PLL}_H
= \mathrm{LSE}_i\!\left(\log \pi_t^{(i)} + \ell^{(i)}_H\right),
\end{equation}
where $\mathrm{LSE}$ denotes the log-sum-exp operation.
Higher is better. PLL is the primary forecasting metric.

\paragraph{Mean squared error (MSE).}
Let the emission model define a conditional distribution
\(
p(x \mid z)
\)
with mean function
\(
\mu_{\mathrm{emit}}(z) := \mathbb{E}[x \mid z].
\)
The predictive mean at horizon $H$ is
\begin{equation}
\mathbb{E}[x_{t+H} \mid x_{1:t}]
= \int \mu_{\mathrm{emit}}(z_{t+H})\,
p(z_{t+H} \mid x_{1:t})\,dz_{t+H},
\end{equation}
where the predictive latent distribution is obtained by propagating the filtering distribution forward,
\begin{equation}
p(z_{t+H} \mid x_{1:t})
= \int p(z_{t+H} \mid z_t)\,p(z_t \mid x_{1:t})\,dz_t.
\end{equation}
The mean squared error is then
\begin{equation}
\mathrm{MSE}_H
= \left\|\mathbb{E}[x_{t+H} \mid x_{1:t}] - x_{t+H}\right\|^2.
\end{equation}

Substituting the filtering approximation \eqref{eq:filt_approx} yields the mixture predictive mean,
\begin{equation}
\mathbb{E}[x_{t+H} \mid x_{1:t}]
\approx \sum_{i=1}^K \pi_t^{(i)}
\int \mu_{\mathrm{emit}}(z_{t+H})\,
p(z_{t+H} \mid z_t^{(i)})\,dz_{t+H}.
\end{equation}

\textit{Numerical evaluation.}
We approximate the inner expectation via Monte Carlo rollouts. For each hypothesis $i$, we draw $M$ samples $z_{t+H}^{(i,m)}$ from the predictive latent dynamics and estimate
\begin{equation}
\hat{x}_{t+H}
= \sum_{i=1}^K \pi_t^{(i)}
\frac{1}{M} \sum_{m=1}^M \mu_{\mathrm{emit}}\!\bigl(z_{t+H}^{(i,m)}\bigr),
\qquad
\mathrm{MSE}_H = \|\hat{x}_{t+H} - x_{t+H}\|^2.
\end{equation}

\paragraph{Predictive branch accuracy (BA).}
Branch accuracy measures the probability that the predictive distribution assigns
mass to the correct latent basin at horizon $H$.
Theoretically,
\begin{equation}
\mathrm{BA}_H
= \mathbb{E}\!\left[
\mathbf{1}\!\left[\mathrm{sign}(z_{t+H}) = \mathrm{sign}(z^\star_{t+H})\right]
\mid x_{1:t}
\right],
\end{equation}
where $z^\star_{t+H}$ is the ground-truth latent.
Substituting \eqref{eq:filt_approx},
\begin{equation}
\mathrm{BA}_H
\approx \sum_{i=1}^K \pi_t^{(i)}
\int \mathbf{1}\!\left[\mathrm{sign}(z_{t+H}) = \mathrm{sign}(z^\star_{t+H})\right]
p(z_{t+1:t+H} \mid z^{(i)}_{1:t})\, dz_{t+1:t+H}.
\end{equation}
$\mathrm{BA}_H = 1$ indicates all predictive mass is in the correct basin;
$\mathrm{BA}_H = 0.5$ is chance.

\textit{Numerical evaluation.}
Using the same $M$ rollouts,
\begin{equation}
b^{(i)}_H
= \frac{1}{M}\sum_{m=1}^M
\mathbf{1}\!\left[
\mathrm{sign}(z^{(i,m)}_{t+H}) = \mathrm{sign}(z^\star_{t+H})
\right],
\qquad
\mathrm{BA}_H = \sum_{i=1}^K \pi_t^{(i)}\, b^{(i)}_H.
\end{equation}

\subsection{Filtering metrics}

Filtering metrics assess the quality of the approximate filtering distribution
\eqref{eq:filt_approx} at time $t$, by comparison against the ground-truth latent
$z^\star_t$.

\paragraph{Latent MSE, bias, and variance.}
The posterior mean latent estimate is
\begin{equation}
\hat{z}_t = \mathbb{E}[z_t \mid x_{1:t}]
= \int z_t\, p(z_t \mid x_{1:t})\, dz_t.
\end{equation}
The MSE decomposes exactly as
\begin{equation}
\mathbb{E}\!\left[(z_t - z^\star_t)^2 \mid x_{1:t}\right]
= \underbrace{(\hat{z}_t - z^\star_t)^2}_{\mathrm{Bias}_{z_{t}}^{2}}
+ \underbrace{\mathbb{E}\!\left[(z_t - \hat{z}_t)^2 \mid x_{1:t}
\right]}_{\mathrm{\sigma}_{z_{t}}^{2}}.
\end{equation}

\textit{Numerical evaluation.}
Substituting \eqref{eq:filt_approx},
\begin{align}
\hat{z}_t                &= \sum_{i=1}^K \pi_t^{(i)} z^{(i)}_t, \\
\mathrm{Bias}_{z_{t}}    &= \hat{z}_t - z^\star_t, \\
\mathrm{\sigma}_{z_{t}}^{2}     &= \sum_{i=1}^K \pi_t^{(i)} (z^{(i)}_t - \hat{z}_t)^2, \\
\mathrm{MSE}_{z_{t}}     &= \sum_{i=1}^K \pi_t^{(i)} (z^{(i)}_t - z^\star_t)^2.
\end{align}

\paragraph{Filtering branch accuracy (BA$_\mathrm{filt}$).}
The probability that the filtering distribution assigns to the correct basin,
\begin{equation}
\mathrm{BA}^\mathrm{filt}_t
= p(\mathrm{sign}(z_t) = \mathrm{sign}(z^\star_t) \mid x_{1:t})
= \int \mathbf{1}\!\left[\mathrm{sign}(z_t) = \mathrm{sign}(z^\star_t)\right]
p(z_t \mid x_{1:t})\, dz_t.
\end{equation}

\textit{Numerical evaluation.}
Substituting \eqref{eq:filt_approx},
\begin{equation}
\mathrm{BA}^\mathrm{filt}_t
\approx \sum_{i=1}^K \pi_t^{(i)}
\mathbf{1}\!\left[\mathrm{sign}(z^{(i)}_t) = \mathrm{sign}(z^\star_t)\right].
\end{equation}

\paragraph{Effective sample size (ESS).}
The ESS quantifies the degeneracy of the hypothesis weights.
For a continuous filtering distribution approximated by a weighted discrete set,
the ESS is defined as the sample size of an equivalent uniform-weight particle set
that would yield the same variance of importance-weighted estimators,
\begin{equation}
\mathrm{ESS}_t
= \frac{\left(\sum_{i=1}^K \pi_t^{(i)}\right)^2}{\sum_{i=1}^K (\pi_t^{(i)})^2}
= \frac{1}{\sum_{i=1}^K (\pi_t^{(i)})^2},
\end{equation}
where the second equality holds because the weights are normalized.
$\mathrm{ESS}_t = K$ when weights are uniform and $\mathrm{ESS}_t = 1$ when all
weight is concentrated on a single hypothesis.

\paragraph{Weight entropy.}
The Shannon entropy of the mixture weights quantifies the spread of probability
mass across the hypothesis set,
\begin{equation}
\mathbb{H}_t = -\sum_{i=1}^K \pi_t^{(i)} \log \pi_t^{(i)},
\end{equation}
ranging from $\mathbb{H}_t = 0$ (all mass on one hypothesis) to
$\mathbb{H}_t = \log K$ (uniform weights).
Weight entropy is complementary to ESS: both measure hypothesis degeneracy, but
entropy is more sensitive to the shape of the full weight distribution whereas ESS
is dominated by the largest weight.

\section{Experimental Parameters}
\label{app:params}

All experiments share the double-well parameter configuration below.
Table~\ref{tab:params} lists all fixed parameters used across the main experiment
and ablations.

\begin{table}[h]
\centering
\caption{Fixed experimental parameters.}
\label{tab:params}
\begin{tabular}{llll}
\toprule
\textbf{Group} & \textbf{Parameter} & \textbf{Symbol} & \textbf{Value} \\
\midrule
\multirow{3}{*}{Latent dynamics}
  & Well minima position  & $a$         & $3.0$ \\
  & Potential scale       & $V_0$         & $0.06$ \\
  & Time step             & $\Delta t$  & $1.0$ \\
  & Diffusion noise       & $\sigma_z$  & $0.05$ \\
\midrule
\multirow{2}{*}{Emission}
  & Emission boundary     & $d$         & $2.0$ \\
  & Emission noise        & $\sigma_x$  & $0.12$ \\
\midrule
\multirow{2}{*}{Initial condition}
  & Prior mean            & $\mu_0$     & $0.0$ \\
  & Prior std             & $\sigma_0$  & $1.0$ \\
\midrule
\multirow{2}{*}{Sequence}
  & Trajectory length     & $T$         & $200$ \\
  & DD detection threshold & $\tau$     & $0.8$ \\
\midrule
\multirow{2}{*}{Dataset}
  & Trajectories per DD bin  & ---      & $100$ \\
  & DD bins                  & ---      & $(30,80),\,(80,140),\,(140,170)$ \\
\midrule
\multirow{3}{*}{Evaluation}
  & Forecast horizons     & $H$         & $\{1, 5, 10\}$ \\
  & MC rollout samples    & $M$         & $20$ \\
  & Inference seeds       & ---         & $\{0, 1, 2\}$ \\
\bottomrule
\end{tabular}
\end{table}

\section{Additional Ablations}
\label{app:ablations}

\subsection{Scoring ablation}
\label{app:ablation:scoring}

\begin{figure}[H]
    \centering
    \includegraphics[width=\linewidth]{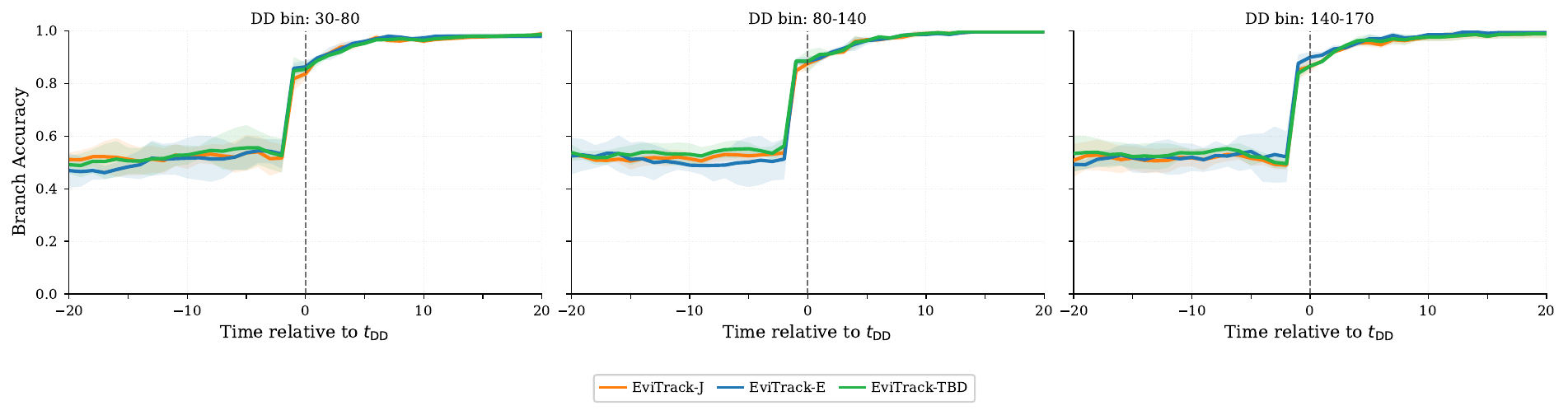}
    \caption{
    \textbf{Scoring function ablation (filtering branch accuracy).}
    Filtering branch accuracy aligned to the true disambiguation time $t_{\mathrm{DD}}$ across DD bins for different scoring rules.
    All variants exhibit similar behavior: accuracy remains near chance before $t_{\mathrm{DD}}$ due to ambiguity and transitions sharply to near-perfect recovery after disambiguation, with only minor differences between scoring functions.
    }
    \label{fig:scoring_ablation_ba}
\end{figure}

\begin{figure}[H]
    \centering
    \includegraphics[width=\linewidth]{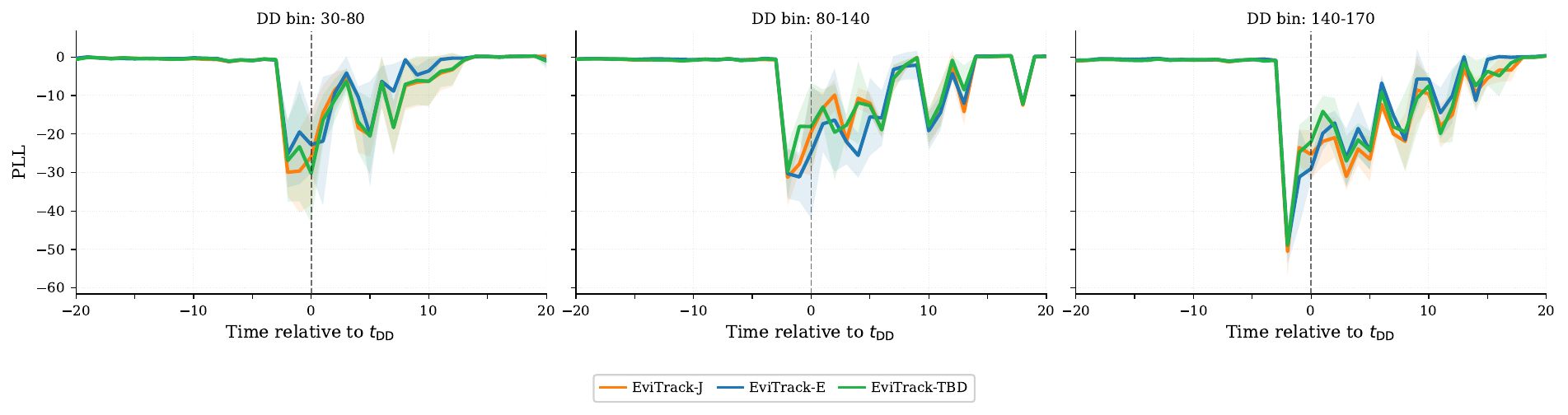}
    \caption{
    \textbf{Scoring function ablation (PLL, $H=1$).}
    Comparison of trajectory scoring rules for EviTrack:
    joint (EviTrack-J), evidence-only (EviTrack-E), and background-normalized (EviTrack-TBD).
    Curves are aligned to the true disambiguation time $t_{\mathrm{DD}}$ and averaged within each DD bin as in Fig.~\ref{fig:main_pll}.
    All variants behave similarly prior to $t_{\mathrm{DD}}$ due to ambiguity.
    After disambiguation, the joint score achieves slightly better predictive performance, indicating that combining prior consistency with observational fit yields improved trajectory ranking.
    }
    \label{fig:scoring_ablation_pll}
\end{figure}

\subsection{Global pruning ablation}
\label{app:ablation:g}

%

\begin{figure}[H]
    \centering
    \includegraphics[width=\linewidth]{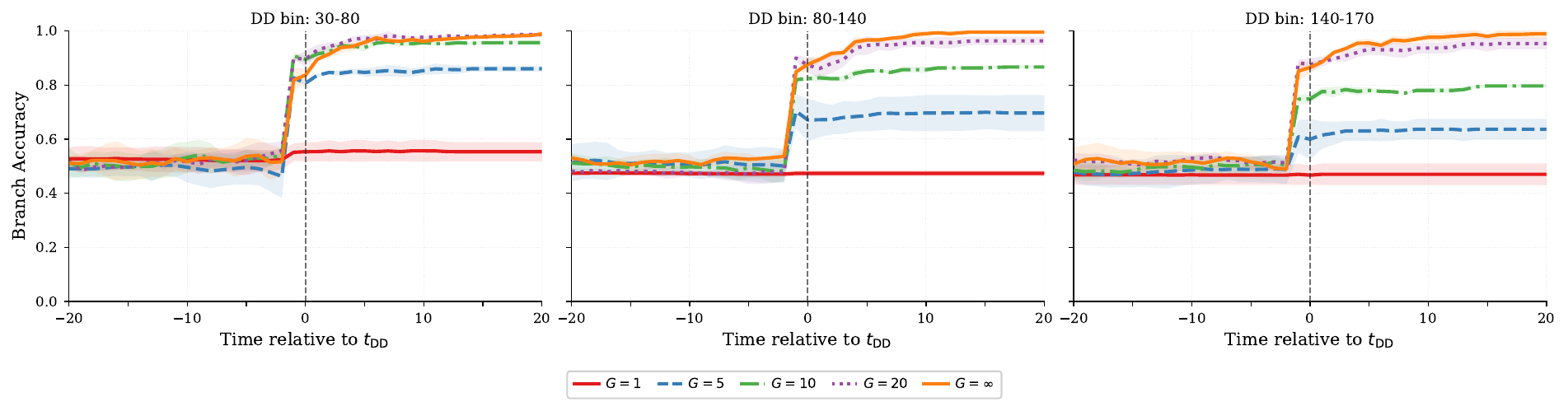}
    \caption{
    \textbf{Global pruning ablation (filtering branch accuracy).}
    Filtering branch accuracy aligned to the true disambiguation time $t_{\mathrm{DD}}$ across DD bins for different global pruning intervals $G$.
    Frequent global pruning ($G=1,5$) prevents recovery of the correct hypothesis after disambiguation, while larger values of $G$ improve accuracy, with $G=\infty$ achieving near-perfect recovery.
    }
    \label{fig:app_g_ablation_ba}
\end{figure}

\begin{figure}[H]
    \centering
    \includegraphics[width=\linewidth]{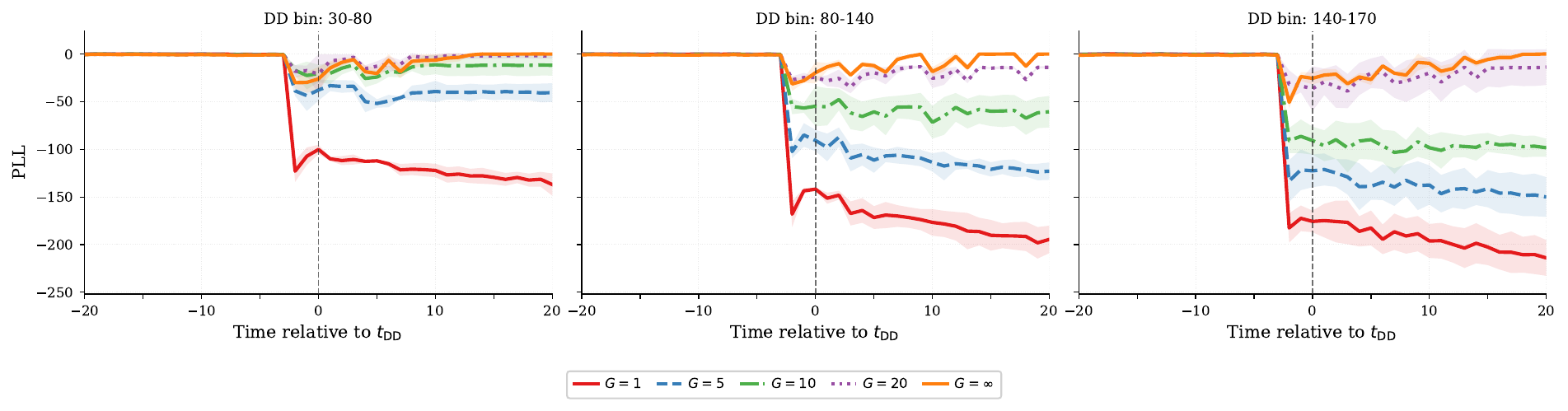}
    \caption{
    \textbf{Global pruning ablation (PLL, $H=1$).}
    Predictive log-likelihood aligned to the true disambiguation time $t_{\mathrm{DD}}$ across DD bins for different global pruning intervals $G$.
    Consistent with the accuracy results in Fig.~\ref{fig:app_g_ablation_ba}, frequent global pruning leads to severe degradation in predictive performance, while preserving hypotheses ($G=\infty$) yields the best results.
    }
    \label{fig:app_g_ablation_pll}
\end{figure}

\subsection{Branching factor ablation}
\label{app:ablation:c}

%

\begin{figure}[H]
    \centering
    \includegraphics[width=\linewidth]{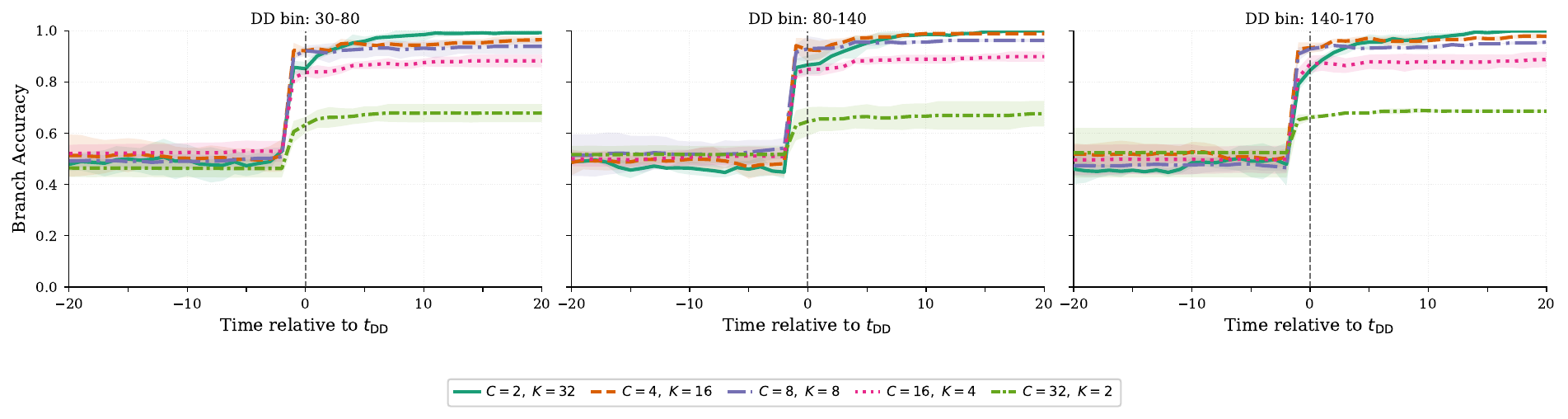}
    \caption{
    \textbf{Branching factor ablation (filtering branch accuracy) at fixed compute ($N=K\cdot C=64$).}
    Filtering branch accuracy aligned to the true disambiguation time $t_{\mathrm{DD}}$ across DD bins for different branching factors $C$.
    Consistent with predictive performance, small branching factors ($C=2,4$) achieve the highest accuracy, while larger values ($C \geq 16$) degrade performance.
    This indicates that excessive local selection reduces hypothesis diversity and limits the ability to recover the correct trajectory after disambiguation.
    }
    \label{fig:c_ablation_ba}
\end{figure}

\begin{figure}[H]
    \centering
    \includegraphics[width=\linewidth]{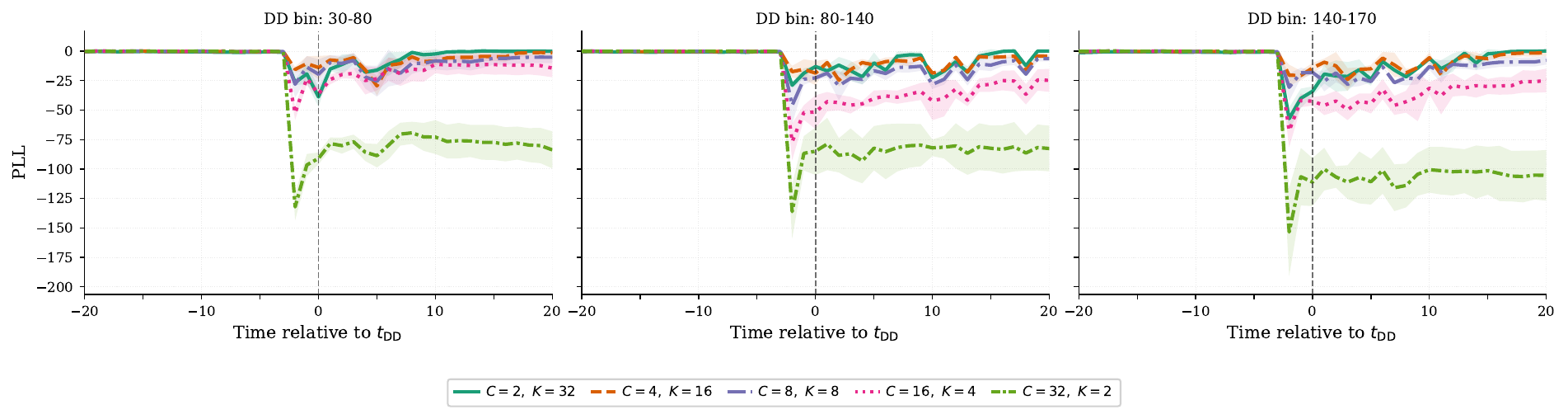}
    \caption{
    \textbf{Branching factor ablation at fixed compute ($N = K \cdot C = 64$).}
    Predictive log-likelihood (PLL, $H=1$) aligned to the true disambiguation time $t_{\mathrm{DD}}$ across DD bins, with curves corresponding to different branching factors $C$ and trajectory counts $K$. Small branching factors ($C=2,4$) achieve the best performance, while larger values ($C \geq 16$) degrade due to overly aggressive local selection among candidate children, which reduces hypothesis diversity prior to disambiguation. This demonstrates that limited selection improves inference, whereas excessive selection harms robustness under delayed disambiguation.
    }
    \label{fig:c_ablation_pll}
\end{figure}

\subsection{Particle count and coverage ablation}
\label{app:ablation:k}

%

\begin{figure}[H]
    \centering
    \includegraphics[width=\linewidth]{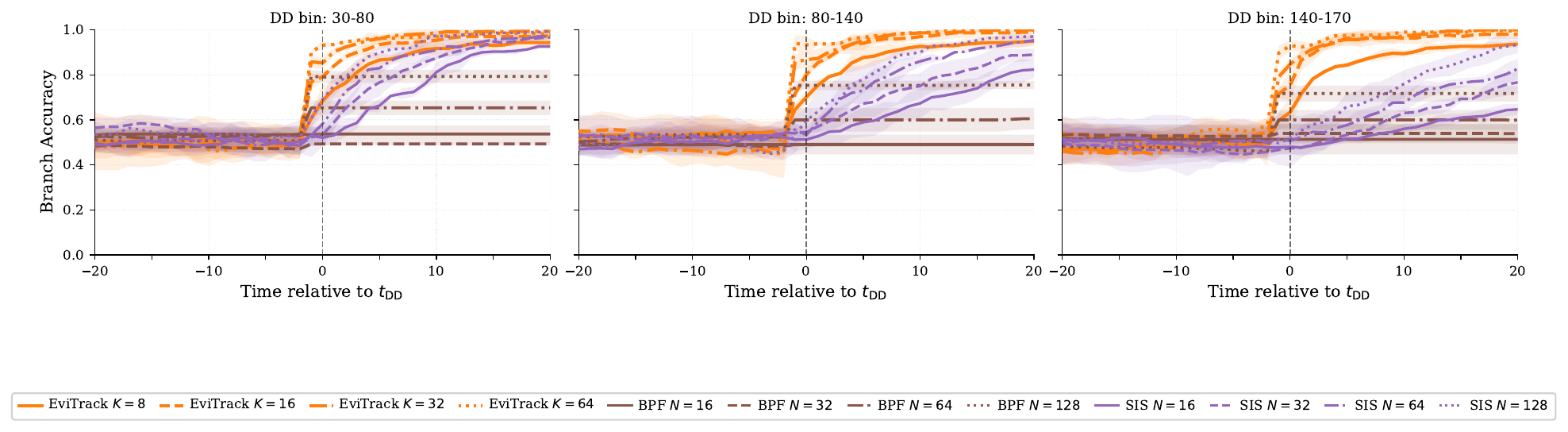}
    \caption{
    \textbf{Particle count ablation (filtering branch accuracy).}
    Filtering branch accuracy aligned to the true disambiguation time $t_{\mathrm{DD}}$ across DD bins for increasing particle counts.
    EviTrack (orange) improves rapidly with moderate $K$, achieving high accuracy shortly after disambiguation.
    In contrast, both Bootstrap PF (brown) and SIS (purple) exhibit substantially lower accuracy even as the number of particles increases.
    This shows that increasing sampling coverage alone does not reliably recover the correct hypothesis in delayed disambiguation settings.
    }
    \label{fig:app_k_ablation_ba}
\end{figure}

\begin{figure}[H]
    \centering
    \includegraphics[width=\linewidth]{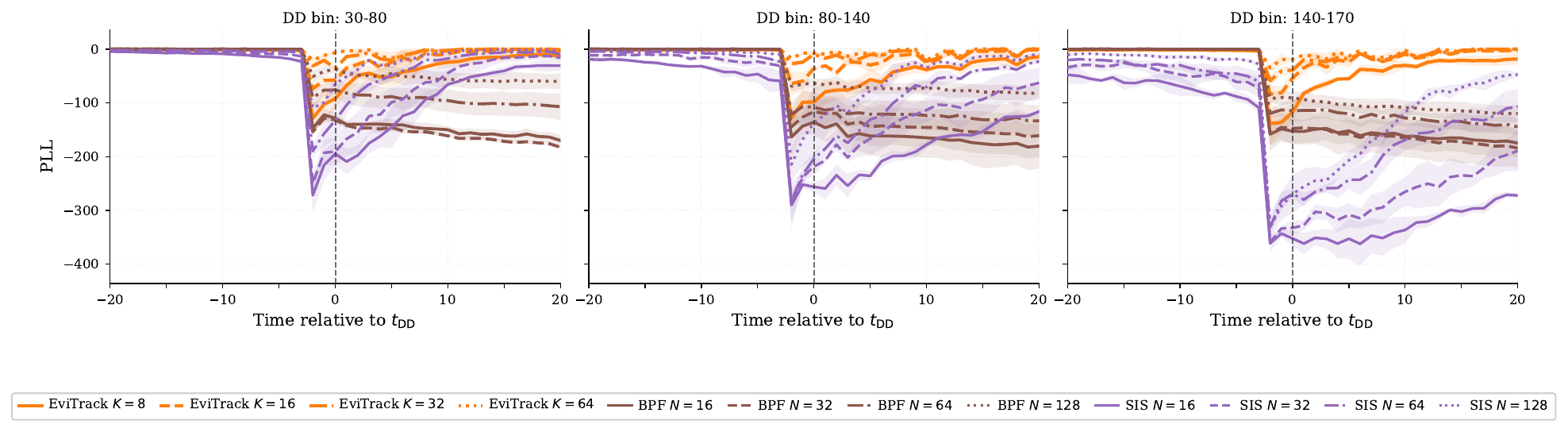}
    \caption{
    \textbf{Particle count ablation (PLL, $H=1$).}
    Predictive log-likelihood aligned to the true disambiguation time $t_{\mathrm{DD}}$ across DD bins for increasing particle counts.
    Consistent with the accuracy results in Fig.~\ref{fig:app_k_ablation_ba}, EviTrack achieves strong predictive performance with relatively small $K$.
    In contrast, Bootstrap PF and SIS remain significantly worse even as the number of particles increases, indicating that brute-force sampling does not close the performance gap.
    }
    \label{fig:app_k_ablation_pll}
\end{figure}

\clearpage

\section{Additional Metrics and Tables}
\label{app:metrics_tables}

\subsection{Main comparison tables}
\label{app:tables:main}

%

\begin{table}[H]
\centering
\caption{
\textbf{Main comparison: filtering branch accuracy before and after disambiguation (all DD bins).}
Values are mean $\pm$ standard deviation across seeds.
}
\label{tab:main_ba_all}
\begin{tabular}{lcc}
\toprule
\textbf{Method} & \textbf{Pre-DD} & \textbf{Post-DD} \\
\midrule
EviTrack-J ($G=\infty$) & $0.512 \pm 0.016$ & $0.987 \pm 0.005$ \\
SIS-PF                  & $0.507 \pm 0.010$ & $0.912 \pm 0.010$ \\
Bootstrap PF            & $0.493 \pm 0.010$ & $0.584 \pm 0.024$ \\
\bottomrule
\end{tabular}
\end{table}

\begin{table}[H]
\centering
\caption{
\textbf{Main comparison: filtering branch accuracy before and after disambiguation by DD bin.}
Values are mean $\pm$ standard deviation across seeds.
}
\label{tab:main_ba_bins}
\begin{tabular}{llcc}
\toprule
\textbf{DD Bin} & \textbf{Method} & \textbf{Pre-DD} & \textbf{Post-DD} \\
\midrule

\multirow{3}{*}{30--80}
& EviTrack-J ($G=\infty$) & $0.518 \pm 0.029$ & $0.991 \pm 0.003$ \\
& SIS-PF                  & $0.514 \pm 0.019$ & $0.984 \pm 0.003$ \\
& Bootstrap PF            & $0.507 \pm 0.021$ & $0.623 \pm 0.025$ \\
\midrule

\multirow{3}{*}{80--140}
& EviTrack-J ($G=\infty$) & $0.514 \pm 0.030$ & $0.993 \pm 0.001$ \\
& SIS-PF                  & $0.520 \pm 0.009$ & $0.946 \pm 0.007$ \\
& Bootstrap PF            & $0.479 \pm 0.005$ & $0.557 \pm 0.034$ \\
\midrule

\multirow{3}{*}{140--170}
& EviTrack-J ($G=\infty$) & $0.504 \pm 0.053$ & $0.978 \pm 0.010$ \\
& SIS-PF                  & $0.489 \pm 0.019$ & $0.806 \pm 0.034$ \\
& Bootstrap PF            & $0.493 \pm 0.016$ & $0.572 \pm 0.069$ \\
\bottomrule
\end{tabular}
\end{table}

\begin{table}[H]
\centering
\caption{
\textbf{Main comparison: predictive log-likelihood (PLL, $H=1$) before and after disambiguation (all DD bins).}
Values are mean $\pm$ standard deviation across seeds.
}
\label{tab:main_pll_all}
\begin{tabular}{lcc}
\toprule
\textbf{Method} & \textbf{Pre-DD} & \textbf{Post-DD} \\
\midrule
EviTrack-J ($G=\infty$) & $-0.784 \pm 0.066$  & $-2.948 \pm 0.103$ \\
SIS-PF                  & $-8.381 \pm 0.118$  & $-54.254 \pm 7.349$ \\
Bootstrap PF            & $-2.419 \pm 0.111$  & $-154.155 \pm 5.832$ \\
\bottomrule
\end{tabular}
\end{table}

\begin{table}[H]
\centering
\caption{
\textbf{Main comparison: predictive log-likelihood (PLL, $H=1$) before and after disambiguation by DD bin.}
Values are mean $\pm$ standard deviation across seeds.
}
\label{tab:main_pll_bins}
\begin{tabular}{llcc}
\toprule
\textbf{DD Bin} & \textbf{Method} & \textbf{Pre-DD} & \textbf{Post-DD} \\
\midrule

\multirow{3}{*}{30--80}
& EviTrack-J ($G=\infty$) & $-1.407 \pm 0.253$ & $-0.851 \pm 0.499$ \\
& SIS-PF                  & $-8.333 \pm 0.923$ & $-5.492 \pm 1.475$ \\
& Bootstrap PF            & $-3.838 \pm 0.288$ & $-136.807 \pm 7.740$ \\
\midrule

\multirow{3}{*}{80--140}
& EviTrack-J ($G=\infty$) & $-0.568 \pm 0.027$ & $-1.563 \pm 0.310$ \\
& SIS-PF                  & $-7.044 \pm 0.123$ & $-26.027 \pm 0.632$ \\
& Bootstrap PF            & $-2.131 \pm 0.372$ & $-167.036 \pm 24.178$ \\
\midrule

\multirow{3}{*}{140--170}
& EviTrack-J ($G=\infty$) & $-0.378 \pm 0.028$ & $-6.429 \pm 0.528$ \\
& SIS-PF                  & $-9.767 \pm 0.458$ & $-131.242 \pm 20.779$ \\
& Bootstrap PF            & $-1.289 \pm 0.269$ & $-158.622 \pm 20.583$ \\
\bottomrule
\end{tabular}
\end{table}

\subsection{Scoring ablation tables}
\label{app:tables:scoring}

%

\begin{table}[H]
\centering
\caption{
\textbf{Scoring ablation: filtering branch accuracy before and after disambiguation (all DD bins).}
Values are mean $\pm$ standard deviation across seeds.
}
\label{tab:scoring_ba_all}
\begin{tabular}{lcc}
\toprule
\textbf{Method} & \textbf{Pre-DD} & \textbf{Post-DD} \\
\midrule
EviTrack-J ($G=\infty$)   & $0.512 \pm 0.016$ & $0.987 \pm 0.005$ \\
EviTrack-E ($G=\infty$)   & $0.507 \pm 0.016$ & $0.990 \pm 0.003$ \\
EviTrack-TBD ($G=\infty$) & $0.516 \pm 0.012$ & $0.988 \pm 0.005$ \\
\bottomrule
\end{tabular}
\end{table}

\begin{table}[H]
\centering
\caption{
\textbf{Scoring ablation: filtering branch accuracy before and after disambiguation by DD bin.}
Values are mean $\pm$ standard deviation across seeds.
}
\label{tab:scoring_ba_bins}
\begin{tabular}{llcc}
\toprule
\textbf{DD Bin} & \textbf{Method} & \textbf{Pre-DD} & \textbf{Post-DD} \\
\midrule

\multirow{3}{*}{30--80}
& EviTrack-J ($G=\infty$)   & $0.518 \pm 0.029$ & $0.991 \pm 0.003$ \\
& EviTrack-E ($G=\infty$)   & $0.502 \pm 0.032$ & $0.991 \pm 0.003$ \\
& EviTrack-TBD ($G=\infty$) & $0.514 \pm 0.022$ & $0.991 \pm 0.004$ \\
\midrule

\multirow{3}{*}{80--140}
& EviTrack-J ($G=\infty$)   & $0.514 \pm 0.030$ & $0.993 \pm 0.001$ \\
& EviTrack-E ($G=\infty$)   & $0.508 \pm 0.037$ & $0.991 \pm 0.005$ \\
& EviTrack-TBD ($G=\infty$) & $0.518 \pm 0.030$ & $0.994 \pm 0.001$ \\
\midrule

\multirow{3}{*}{140--170}
& EviTrack-J ($G=\infty$)   & $0.504 \pm 0.053$ & $0.978 \pm 0.010$ \\
& EviTrack-E ($G=\infty$)   & $0.511 \pm 0.009$ & $0.987 \pm 0.004$ \\
& EviTrack-TBD ($G=\infty$) & $0.516 \pm 0.049$ & $0.980 \pm 0.012$ \\
\bottomrule
\end{tabular}
\end{table}

\begin{table}[H]
\centering
\caption{
\textbf{Scoring ablation: predictive log-likelihood (PLL, $H=1$) before and after disambiguation (all DD bins).}
Values are mean $\pm$ standard deviation across seeds.
}
\label{tab:scoring_pll_all}
\begin{tabular}{lcc}
\toprule
\textbf{Method} & \textbf{Pre-DD} & \textbf{Post-DD} \\
\midrule
EviTrack-J  ($G=\infty$)   & $-0.784 \pm 0.066$ & $-2.948 \pm 0.103$ \\
EviTrack-E  ($G=\infty$)   & $-0.614 \pm 0.116$ & $-2.459 \pm 0.100$ \\
EviTrack-TBD ($G=\infty$) & $-0.673 \pm 0.186$ & $-2.631 \pm 0.400$ \\
\bottomrule
\end{tabular}
\end{table}

\begin{table}[H]
\centering
\caption{
\textbf{Scoring ablation: predictive log-likelihood (PLL, $H=1$) before and after disambiguation by DD bin.}
Values are mean $\pm$ standard deviation across seeds.
}
\label{tab:scoring_pll_bins}
\begin{tabular}{llcc}
\toprule
\textbf{DD Bin} & \textbf{Method} & \textbf{Pre-DD} & \textbf{Post-DD} \\
\midrule

\multirow{3}{*}{30--80}
& EviTrack-J ($G=\infty$)   & $-1.407 \pm 0.253$ & $-0.851 \pm 0.499$ \\
& EviTrack-E ($G=\infty$)   & $-1.031 \pm 0.373$ & $-0.588 \pm 0.232$ \\
& EviTrack-TBD ($G=\infty$) & $-1.187 \pm 0.418$ & $-0.928 \pm 0.388$ \\
\midrule

\multirow{3}{*}{80--140}
& EviTrack-J ($G=\infty$)   & $-0.568 \pm 0.027$ & $-1.563 \pm 0.310$ \\
& EviTrack-E ($G=\infty$)   & $-0.510 \pm 0.092$ & $-1.909 \pm 0.410$ \\
& EviTrack-TBD ($G=\infty$) & $-0.454 \pm 0.108$ & $-1.548 \pm 0.179$ \\
\midrule

\multirow{3}{*}{140--170}
& EviTrack-J ($G=\infty$)   & $-0.378 \pm 0.028$ & $-6.429 \pm 0.528$ \\
& EviTrack-E ($G=\infty$)   & $-0.299 \pm 0.112$ & $-4.881 \pm 0.561$ \\
& EviTrack-TBD ($G=\infty$) & $-0.376 \pm 0.039$ & $-5.416 \pm 1.270$ \\
\bottomrule
\end{tabular}
\end{table}



\end{document}